\documentclass[final]{cvpr}

\usepackage{times}
\usepackage{epsfig}
\usepackage{graphicx}
\usepackage{amsmath}
\usepackage{amssymb}
\usepackage{microtype}
\usepackage{multirow}
\usepackage{subfig}

\usepackage[pagebackref=true,breaklinks=true,colorlinks,bookmarks=false]{hyperref}

\def\cvprPaperID{****} % *** Enter the CVPR Paper ID here
\def\confYear{CVPR 2021}

\begin{document}

%%%%%%%%% TITLE
\title{Two-Stream Consensus Network: Submission to HACS Challenge 2021 Weakly-Supervised Learning Track}

\author{Yuanhao Zhai$^{1,2}$ \quad Le Wang$^{1}$ \quad David Doermann$^{2}$ \quad Junsong Yuan$^{2}$ \\
$^1$Xi'an Jiaotong University \\
$^1$State University of New York at Buffalo \\
{\tt\small yzhai6@buffalo.edu}
% {\tt\small lewang@xjtu.edu.cn} \\
% {\tt\small \{yzhai6, doermann, jsyuan\}@buffalo.edu}
}

\maketitle

%%%%%%%%% ABSTRACT
\begin{abstract}
This technical report presents our solution to the HACS Temporal Action Localization Challenge 2021, Weakly-Supervised Learning Track.
The goal of weakly-supervised temporal action localization is to temporally locate and classify action of interest in untrimmed videos given only video-level labels.
We adopt the two-stream consensus network~(TSCN)~\cite{zhai2020two} as the main framework in this challenge.
The TSCN consists of a two-stream base model training procedure and a pseudo ground truth learning procedure.
The base model training encourages the model to predict reliable predictions based on single modality (\ie{}, RGB or optical flow), based on the fusion of which a pseudo ground truth is generated and in turn used as supervision to train the base models.
On the HACS v1.1.1 dataset, without fine-tuning the feature-extraction I3D models, our method achieves $22.20\%$ on the validation set and $21.68\%$ on the testing set in terms of average mAP.
Our solution ranked the 2nd in this challenge, and we hope our method can serve as a baseline for future academic research.
\end{abstract}

\section{Our Solution}
\begin{figure*}
	\centering
	\includegraphics[width=0.85\linewidth]{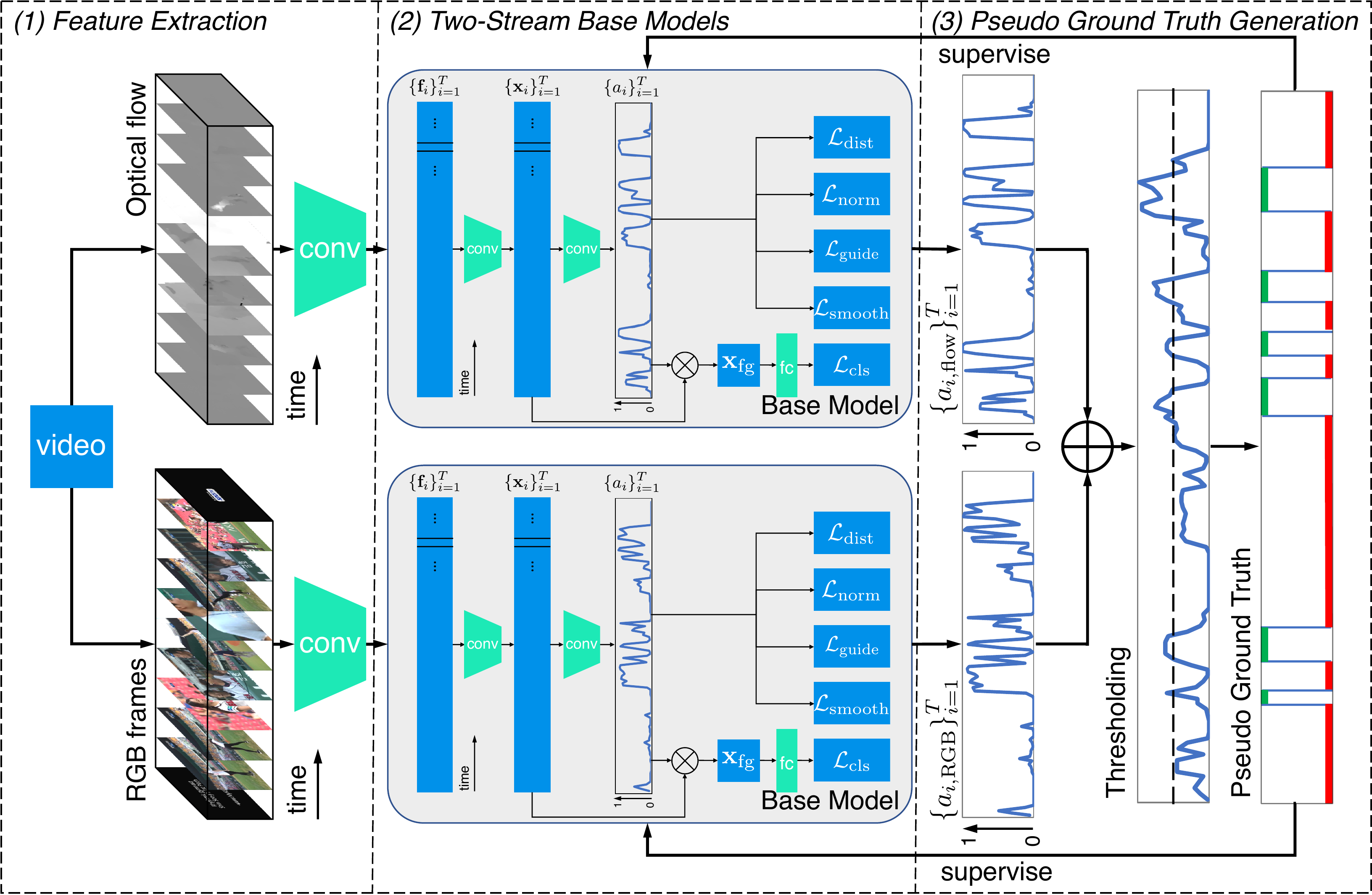}
	\caption{An overview of the Two-Stream Consensus Network, which consists of three parts: (1) feature extraction, where RGB and optical flow snippet-level features are pre-extracted; (2) two-stream base models, where action recognition is performed on the two modalities with two-stream base models, respectively; (3) pseudo ground truth generation, where a frame-level pseudo ground truth is generated from the two-stream late-fusion attention sequence and in turn provides frame-level supervision to two-stream base models.}
	\label{fig:framework}
\end{figure*}

We adopt the two-stream consensus network (TSCN)~\cite{zhai2020two} as the main framework.
It consists of two main procedures: two-stream base model training and pseudo ground truth learning.
\figurename~\ref{fig:framework} shows the framework of our method.

\subsection{Feature Extraction}
We construct our model upon snippet-level feature sequences extracted from the raw video volume.
The RGB and optical flow features are extracted with pre-trained I3D~\cite{carreira2017quo}) from non-overlapping fixed-length RGB frame snippets and optical flow snippets, respectively.
Formally, given a video with $T$ non-overlapping snippets, we denote the RGB features and optical flow features as $\{ \mathbf{f}_{\text{RGB},i} \}_{i=1}^T$ and $\{ \mathbf{f}_{\text{flow},i} \}_{i=1}^T$, respectively, where $\mathbf{f}_{\text{RGB},i}, \mathbf{f}_{\text{flow},i} \in \mathbb{R}^{D}$ are the feature representations of the $i$-th RGB frame and optical flow snippet, respectively, and $D$ represents the channel dimension.

\subsection{Two-Stream Base Models}
After obtaining the RGB and optical flow features, we first use two-stream base models to perform the video-level action classification.
The features of two modalities are fed into two separate base models, respectively, and the two base models use the same architecture but do not share parameters.
Therefore, we omit the subscript $\text{RGB}$ and $\text{flow}$ to denote a general operation for both modalities.

To embed the extracted features to task-specific space, we use a single temporal convolutional layer with a kernel size $3$ to embed the input feature, and generate a set of new features $\{ \mathbf{x}_i \}_{i=1}^{T}$, where $\mathbf{x}_i \in \mathbb{R}^{D}$.

As a video may contain background snippets, to perform video-level classification, we need to select snippets that are likely to contain action instances and meanwhile filter out snippets that are likely to contain background. To this end, an attention value $a_i \in (0, 1)$ to measure the likelihood of the $i$-th snippet containing an action is given by a fully-connected (FC) sigmoid layer.
We then perform attention-weighted pooling over the feature sequence to generate a single foreground feature $\mathbf{x}_{\text{fg}}$, and feed it to an FC softmax layer to get the video-level prediction:
\begin{equation}
	\mathbf{x}_{\text{fg}} = \frac{1}{\sum_{i=1}^{T} a_{i}} \sum_{i=1}^{T}a_{i} \mathbf{x}_i,
\end{equation}
\begin{equation}
	\hat{y}_{c} = \frac{e^{\mathbf{w}_c \cdot \mathbf{x}_{\text{fg}} + b_c}}{\sum_{i=1}^{C}e^{\mathbf{w}_i \cdot \mathbf{x}_{\text{fg}} + b_i}},
\end{equation}
where $\hat{y}_{c}$ is the probability that the video contains the $c$-th action, and $\mathbf{w}_c$ and $b_c$ are the weight and bias of the FC layer for category $c$, respectively.
The classification loss function $\mathcal{L}_{\text{cls}}$ is defined as the standard binary cross entropy loss.

% \begin{equation}
% 	\mathcal{L}_{\text{cls}} = -\sum_{c=1}^{C} [y_{c}\log (\hat{y}_{c}) + (1 - y_c)\log(1 - \hat{y}_c)],
% \end{equation}
% where $y_{c}$ denotes the value of label vector $\mathbf{y}$ at index $c$.
% The label vector $\mathbf{y}$ is a multi-hot vector, and its $c$-th value is set to $1$ if the video contains actions that belongs to the $c$-th category, otherwise $0$.

In addition, the temporal-class activation map (T-CAM) $\{ \textbf{s}_i \}_{i=1}^{T}$, $\textbf{s}_i \in \mathbb{R}^{C}$ is also generated by sliding the classification FC softmax layer over all snippet features: 
\begin{equation}
	s_{i,c} = \frac{e^{\mathbf{w}_c \cdot \mathbf{x}_{i} + b_c}}{\sum_{j=1}^{C}e^{\mathbf{w}_j \cdot \mathbf{x}_{i} + b_j}},
\end{equation}
where $s_{i,c}$ is the T-CAM value of $i$-th snippet for category $c$.

For better attention and T-CAM learning, we further adopt a smooth loss, the attention normalization loss~\cite{zhai2020two}, a distinctness loss, and a variant of the self-guided attention loss~\cite{nguyen2019weakly} in the base model training.

The smooth loss enforces temporally proximate snippets to give similar attention predictions, and thus helps generate a more smooth attention sequence:
\begin{equation}
    \mathcal{L}_{\text{smooth}} = \frac{1}{T-1}  \sum_{t=1}^{T-1} |a_t - a_{t+1}| 
\end{equation}

The attention normalization loss~\cite{zhai2020two} maximizes the difference between the average top-$l$ attention values and the average bottom-$l$ attention values, and forces the foreground attention to be $1$ and background attention to be $0$:
\begin{equation}
	\mathcal{L}_{\text{norm}} = \frac{1}{l}\min_{\substack{a \subset \{a_i\} \\|a|=l}} \sum_{\phi \in a}\phi - \frac{1}{l} \max_{\substack{a \subset \{ai\} \\|a|=l}} \sum_{\phi \in a}\phi,
	\label{eq:lNorm}
\end{equation}
where $l=\max \left(1, \lfloor\frac{T}{k}\rfloor \right)$ and $k$ is a hyperparameter to control the selected snippets.

The distinctness loss $\mathcal{L}_{\text{dist}}$ encourages the foreground feature $\mathbf{x}_{\text{fg}}$ and background feature $\mathbf{x}_{\text{bg}}$ to be distinct in the feature space:
\begin{equation}
	\mathcal{L}_{\text{dist}} = \max \left ( 0, \frac{\mathbf{x}_{\text{fg}} \cdot \mathbf{x}_{\text{bg}}}{\| \mathbf{x}_{\text{fg}} \| \| \mathbf{x}_{\text{bg}} \|} - m   \right ),
\end{equation}
where $\| \cdot \|$ is the L2 norm, and $m$ is a hyperparameter empirically set to $0.5$.

The self-guided attention loss~\cite{nguyen2019weakly} pursues a consensus between the bottom-up attention and the top-down T-CAM.
In our method, as we do not exploit the background classification, we discard the background modeling term in the original guide loss.
Besides, we empirically observe that the attention tends to produce more reliable activations than the T-CAM, and thus we detach the gradient for the attention in the guide loss.
The guide loss variant employed in our method is formulated as:
\begin{equation}
    \mathcal{L}_{\text{guide}} = \frac{1}{T} \sum_{i=1}^{T} | \text{sg}(a_i) - s_{i, c^{*}}|,
\end{equation}
where $\text{sg}(\cdot)$ denotes stop gradient, and $c^{*}$ is the ground truth action class\footnote{If there are multiple classes contained in the video, we max-pool the T-CAM across all ground truth classes.}.

The overall loss for the base model training is a weighted sum of the loss terms:
\begin{equation}
	\mathcal{L}_{\text{base}} = \mathcal{L}_{\text{cls}} + \lambda_1 \mathcal{L}_{\text{smooth}} + \lambda_2 \mathcal{L}_{\text{norm}} + \lambda_3 \mathcal{L}_{\text{dist}} + \lambda_4 \mathcal{L}_{\text{guide}},
\end{equation}
where $\lambda_1$, $\lambda_2$, $\lambda_3$ and $\lambda_4$ are weight parameters.

\subsection{Pseudo Ground Truth Learning}
We iteratively refine the two-stream base models with a frame-level pseudo ground truth, which is generated by two-stream prediction fusion.
Specifically, we divide the whole training process into several refinement iterations. 
At refinement iteration $0$, only video-level labels are used for training.
And at refinement iteration $n+1$, a frame-level pseudo ground truth is generated at refinement iteration $n$, and provides frame-level supervision for the current refinement iteration.

Speficially, we use the fusion attention sequence $\{ a_{\text{fuse},i}^{(n)} \}_{i=1}^T$ at refinement iteration $n$ to generate pseudo ground truth $\{ \mathcal{G}_{i}^{(n+1)} \}_{i=1}^T$ for refinement iteration $n + 1$, where $a_{\text{fuse},i}^{(n)}=\beta a_{\text{RGB},i}^{(n)} + (1-\beta)a_{\text{flow},i}^{(n)}$, and $\beta \in [0, 1]$ is a hyperparameter to control the relative importance of RGB and flow attentions.

The pseudo ground truth thresholds the attention sequence to generate a binary sequence:
\begin{equation}
	\mathcal{G}_{i}^{(n+1)} = 
	\left \{
	\begin{array}{lr}
		1, \quad a_{\text{fuse},i}^{(n)} > \theta; & \\
		0, \quad a_{\text{fuse},i}^{(n)} \leq \theta, & \\
	\end{array}
	\right.
\end{equation}
where $\theta$ is the threshold value.
% Setting a large value of $\theta$ will eliminate the action proposals that only one stream has high activations, and therefore reduces the false positive rate. In contrast, setting a small value of $\theta$ will help models to generate more action proposals and achieve a higher recall. Hard pseudo labels remove the uncertainty and provide stronger supervision, but introduce a hyperparameter.

After obtaining the frame-level pseudo ground truth, we force the attention sequence generated by \textit{each} stream to be similar to the pseudo ground truth with a binary cross entropy loss:
\begin{equation}
	\begin{aligned}
		\mathcal{L}_{\text{pseudo}}^{(n + 1)} = & -\frac{1}{T} \sum_{i=1}^{T} \mathcal{G}_{i}^{(n + 1)} \log \left ( a_{i}^{(n + 1)} \right ) + \\
		& \left ( 1 - \mathcal{G}_{i}^{(n + 1)} \right ) \log \left ( 1 - a_{i}^{(n + 1)} \right )
	\end{aligned}
\end{equation}
At refinement iteration $n + 1$, the total loss for each stream is 
\begin{equation}
	\mathcal{L}_{\text{total}}^{(n + 1)} = \mathcal{L}_{\text{cls}} + \lambda_3 \mathcal{L}_{\text{dist}} + \lambda_4 \mathcal{L}_{\text{guide}}, + \lambda_5 \mathcal{L}_{\text{pseudo}}^{(n + 1)},
\end{equation}
where $\lambda_5$ is a hyperparameter to control the relative importance of two losses.
Note that we remove the attention normalization loss and in the pseudo ground truth learning, as the loss term assumes at least $\frac{1}{k}$ and $\frac{1}{k}$ of each video are actions and backgrounds, respectively.
However, it does not always hold (some videos do not contain background), and will bring noise for the pseudo ground truth learning.
The smooth loss is also removed as it leads to trivial solution where all videos are actions without the supervision the attention normalization loss.

\subsection{Action Localization}
During testing, we first temporally upsample the attention sequence and T-CAM by a factor of $8$ via linear interpolation.
Then, we select top-$2$ action categories from the fusion video-level prediction $\hat{\textbf{y}}_{\text{fuse}}$ to perform action localization, where $\hat{\textbf{y}}_{\text{fuse}}=\beta \hat{\textbf{y}}_{\text{RGB}} + (1-\beta) \hat{\textbf{y}}_{\text{flow}}$.
% For each of these categories, following our intention that the attention performs a binary selection, we generate action proposals by directly thresholding the attention value at $0.5$ and concatenating consecutive snippets.
Action proposals are generated by progressively thresholding the attention sequence from $0$ to $1.0$, with a step size of $0.025$, and concatenating proximate snippets.
The action proposals are scored following TSCN~\cite{zhai2020two}.
Formally, given action proposal $(t_s, t_e, c)$, fusion attention $\{a_{\text{fuse},i} \}_{i=1}^{T}$ and T-CAM $\{ \textbf{s}_{\text{fuse},i} \}_{i=1}^{T}$, where $\textbf{s}_{\text{fuse},i} = \beta \textbf{s}_{\text{RGB},i} + (1 - \beta)\textbf{s}_{\text{flow},i} $, the score $\psi$ is computed as 
\begin{equation}
     \begin{aligned}
        \psi = & \frac{\sum_{i=t_s}^{t_e} a_{\text{fuse},i} s_{\text{fuse},i,c}}{t_e - t_s}
        - \\
        & \frac{\sum_{i=T_s}^{T_e} a_{\text{fuse},i} s_{\text{fuse},i,c} - \sum_{i=t_s}^{t_e} a_{\text{fuse},i} s_{\text{fuse},i,c}}{T_e - T_s - (t_e - t_s)}, 
     \end{aligned}
\end{equation}
where $T_s=t_s-\frac{L}{4}$, $T_e=t_e+\frac{L}{4}$, and $L=t_e-t_s$.
We finally use NMS with IoU threshold $0.6$ to filter out redundant detections.

% \subsection{Pseudo Ground Truth Learning}

\section{Experiments and Discussions}

\subsection{Implementation Details}
The optical flow is estimated via the TV-L1 algorithm~\cite{zach2007duality}.
The feature-extraction backbone I3D~\cite{carreira2017quo} is pre-trained on the Kinetics dataset~\cite{carreira2017quo}, and is \emph{not} fine-tuned on the HACS dataset~\cite{zhao2019hacs}.
In this competition, we use the off-the-shelf $2$ FPS RGB snippet-level features provided by the dataset~\cite{zhao2019hacs}, and extract the optical flow features with a snippet length of $16$ frames.
The majority of the hyperparameters are set according to \cite{nguyen2019weakly,zhai2020two}: $\lambda_1=\lambda_2=\lambda_4=0.1$, $k=8$, and $\theta=0.5$.
Other hyperparameters are set according to a grid search: $\beta=0.6$, $\lambda_3=0.1$ and $\lambda_5=0.01$.
We use the AdamW optimizer with a fixed learning rate $0.0005$.
We train the model for a total of $5$ refinement iterations, with each refinement iteration contains $10$ epochs.
At each refinement iteration, we simply select the latest model from the last refinement iteration to generate the pseudo ground truth.

\newlength{\savecw}
\begin{table*}[t!]
\centering
\small
\setlength{\savecw}{\columnwidth}
\caption{Single-stream performance with different loss combinations on the HACS validation set.}\label{tab:single-stream}

\subfloat[%
  RGB stream-only localization performance. \label{subtab:flow}
]{%
  \begin{minipage}{\savecw}
  \centering
	\begin{tabular}{ccccc|ccc|c}
		\hline
		\multirow{2}{*}{$\mathcal{L}_{\text{cls}}$} & 
		\multirow{2}{*}{$\mathcal{L}_{\text{smooth}}$} & \multirow{2}{*}{$\mathcal{L}_{\text{norm}}$} & \multirow{2}{*}{$\mathcal{L}_{\text{dist}}$} & \multirow{2}{*}{$\mathcal{L}_{\text{guide}}$} & \multicolumn{4}{c}{mAP@IoU (\%)} \\
		& & & & & 0.5 & 0.75 & 0.95 & Avg \\
		\hline
		\hline
		\checkmark & - & - & - & - & 9.23 & 4.45 & 1.12 & 5.03 \\ 
		\checkmark & \checkmark & - & - & - & 17.53 & 11.41 & 4.54 & 11.69 \\ 
		\checkmark & \checkmark & \checkmark & - & - & 25.82 & 15.77 & 5.42 & 16.33 \\ 
		\checkmark & \checkmark & \checkmark & \checkmark & - & 26.54 & 16.18 & 5.50 & 16.86 \\ 
		\checkmark & \checkmark & \checkmark & \checkmark & \checkmark & 27.46 & 17.19 & 6.26 & 17.69 \\ 
		\hline
	\end{tabular}
  \end{minipage}%
}% end of subfloat
% \hfill
\vfill
\subfloat[%
  Flow stream-only localization performance. \label{subtab:rgb}
]{%
  \begin{minipage}{\savecw}
  \centering
  \begin{tabular}{ccccc|ccc|c}
	\hline
	\multirow{2}{*}{$\mathcal{L}_{\text{cls}}$} & 
	\multirow{2}{*}{$\mathcal{L}_{\text{smooth}}$} & \multirow{2}{*}{$\mathcal{L}_{\text{norm}}$} & \multirow{2}{*}{$\mathcal{L}_{\text{dist}}$} & \multirow{2}{*}{$\mathcal{L}_{\text{guide}}$} & \multicolumn{4}{c}{mAP@IoU (\%)} \\
	& & & & & 0.5 & 0.75 & 0.95 & Avg \\
	\hline
	\hline
	\checkmark & - & - & - & - & 5.60 & 2.17 & 0.19 & 2.63 \\ 
	\checkmark & \checkmark & - & - & - & 14.47 & 9.43 & 3.40 & 9.64 \\ 
	\checkmark & \checkmark & \checkmark & - & - & 19.08 & 11.73 & 4.75 & 12.31 \\ 
	\checkmark & \checkmark & \checkmark & \checkmark & - & 19.24 & 11.84 & 4.83 & 12.40 \\ 
	\checkmark & \checkmark & \checkmark & \checkmark & \checkmark & 19.57 & 12.17 & 4.90 & 12.74 \\ 
	\hline
\end{tabular}
  \end{minipage}%
}% end of subfloat
\end{table*}

\begin{table}[t]
	\centering
	\small
	\caption{Performance w/ pseudo ground truth learning in different refinement iterations on the HACS validation and testing sets.}
	\label{tab:comparison-on-pseudo-ground-truth}
	\begin{tabular}{c|c|ccc|c|c}
		\hline
		& \multirow{2}{*}{Modality} & \multicolumn{4}{c|}{Validation} & Test \\
		& & 0.5 & 0.75 & 0.95 & Avg & Avg \\
		\hline
		\hline
		\multirow{3}{*}{0} & RGB & 27.46 & 17.19 & 6.26 & 17.69 & - \\
		& Flow & 19.57 & 12.17 & 4.90 & 12.74 & - \\
		& Fusion & 29.91 & 18.50 & 6.99 & 19.12 & 18.75 \\
		\hline
		\multirow{3}{*}{1} & RGB & 26.85 & 16.98 & 6.27 & 17.38 & - \\
		& Flow & 23.57 & 13.86 & 5.27 & 14.74 & - \\
		& Fusion & 31.30 & 19.14 & 6.96 & 19.83 & - \\
		\hline
		\multirow{3}{*}{2} & RGB & 30.45 & 18.42 & 6.39 & 19.16 & - \\
		& Flow & 24.09 & 14.39 & 5.43 & 15.16 & - \\
		& Fusion & 33.47 & 20.03 & 6.81 & 20.91 & 20.30 \\
		\hline
		\multirow{3}{*}{3} & RGB & 32.62 & 19.32 & 6.36 & 20.18 & - \\
		& Flow & 23.21 & 13.97 & 4.82 & 14.58 & - \\
		& Fusion & 35.07 & 20.85 & 6.87 & 21.82 & - \\
		\hline
		\multirow{3}{*}{4} & RGB & 33.54 & 19.57 & 6.17 & 20.54 & - \\
		& Flow & 23.12 & 13.74 & 4.50 & 14.38 & - \\
		& Fusion & 35.43 & 20.94 & 6.80 & 21.93  & 21.33 \\
		\hline
	\end{tabular}
\end{table}

\subsection{Results}
\noindent\textbf{Two-stream base models}.
The performance of two-stream base models w/o pseudo ground truth supervision is reported in \tablename~\ref{tab:single-stream}, where different combinations of loss terms are evaluated.
The results show the addition of each loss contributes to the performance improvement.

\noindent\textbf{Pseudo ground truth learning}.
\tablename~\ref{tab:comparison-on-pseudo-ground-truth} reports the performance changes in different refinement iterations.
The results reveal that pseudo ground truth consistently improves the fusion results, and eventually saturates at the $3$-rd and $4$-th refinement iterations.
The pseudo ground truth also greatly improves the single-stream models.
Specifically, it improves the performance of the RGB model from $17.69\%$ to $20.54\%$ in terms of average mAP, and improves the performance of the flow model from $12.74\%$ to $15.16\%$.

\noindent\textbf{Exponential moving average emsemble}.
Inspired by the mean teacher~\cite{tarvainen2017mean}, after the pseudo ground truth learning, we ensemble the models from all $5$ refinement iterations by exponentially moving average their parameters with a successive weight $0.2$.
The final ensemble model achieves $22.20\%$ average mAP on the HACS validation set, and $21.68\%$ on the HACS testing set.

% \subsection{Future Directions}

% \noindent\textbf{Multi-modality fusion}.

% \noindent\textbf{Fine-tuning backbone}.

\section*{Acknowledgements}
This work is supported in part by the Defense Advanced Research Projects Agency
(DARPA) under Contract No. HR001120C0124. Any opinions, findings and conclusions
or recommendations expressed in this material are those of the author(s) and do
not necessarily reflect the views of the Defense Advanced Research Projects
Agency (DARPA).

{\small
\bibliographystyle{ieee_fullname}
\bibliography{ref}
}

\end{document}